\documentclass[conference,a4paper]{IEEEtran}
\IEEEoverridecommandlockouts

\usepackage{graphicx}
\usepackage{subfigure}
\usepackage{siunitx}
\usepackage{xcolor}
\usepackage{algorithm}
\usepackage{comment}
\usepackage{varwidth}
\usepackage{subfig} 
\usepackage{algpseudocode}
\usepackage{enumitem}
\usepackage{caption}
\usepackage{balance}
\usepackage[disable]{todonotes}
\usepackage[backend = bibtex,
			style = numeric,
			sorting = none,
			]{biblatex}
\bibliography{references.bib, caleb-zotero.bib}
\renewcommand*{\bibfont}{\small}
\usepackage{url}
\usepackage{multirow}

\newcommand{\eg}[1]{}
\renewcommand{\eg}[1]{(e.g. {#1})}

\newcommand{\ie}[1]{}
\renewcommand{\ie}[1]{(i.e. {#1})}

\newcolumntype{L}[1]{>{\raggedright\let\newline\\\arraybackslash\hspace{0pt}}m{#1}}
\newcolumntype{C}[1]{>{\centering\let\newline\\\arraybackslash\hspace{0pt}}m{#1}}

\begin{document}

\title{Federated Learning in IoT: a Survey from a Resource-Constrained Perspective}

\author{\IEEEauthorblockN{Ishmeet Kaur}
\IEEEauthorblockA{\textit{*Corresponding Author: ishmeet3kk@gmail.com}} 

\and
\IEEEauthorblockN{Adwaita Janardhan Jadhav}
\IEEEauthorblockA{\textit{adwaitas28@gmail.com} }

}

\maketitle
\begin{abstract}

The IoT ecosystem is able to leverage vast amounts of data for intelligent decision-making. Federated Learning (FL), a decentralized machine learning technique, is widely used to collect and train machine learning models from a variety of distributed data sources.  Both IoT and FL systems can be complimentary and used together. 
However, the resource-constrained nature of IoT devices prevents the widescale deployment FL in the real world.
This research paper presents a comprehensive survey of the challenges and solutions associated with implementing Federated Learning (FL) in resource-constrained Internet of Things (IoT) environments, viewed from 2 levels, client and server. 
We focus on solutions regarding limited client resources, presence of heterogeneous client data, server capacity, and high communication costs, and assess their effectiveness in various scenarios. Furthermore, we categorize the solutions based on the location of their application, i.e., the IoT client, and the FL server.
In addition to a comprehensive review of existing research and potential future directions, this paper also presents new evaluation metrics that would allow researchers to evaluate their solutions on resource-constrained IoT devices.

\end{abstract}

\begin{small}
\textbf{\textit{keywords} -- federated learning, internet-of-things, survey}
\end{small}

\section{Introduction}


Internet of Things (IoT) refers to the vast network of interconnected devices embedded with sensors, aimed at exchanging data with each other and the cloud over the internet~\cite{marculescu2020edge}.
In the era of technological advancements, the rapid growth of IoT stands as a key influencer across diverse sectors, ranging from healthcare, automation, and transportation~\cite{lai2021oort}.  

These IoT devices continuously generate a vast amount of data, and to efficiently utilize this rich data source, recently Machine Learning (ML) techniques have been employed~\cite{lai2021oort}. In many of these applications, ML models are primarily trained at a centralized location, such as a cloud server, where data originating from various IoT devices is aggregated~\cite{mcmahan2017communication}. These trained models are then deployed to the IoT devices where they perform tasks like object classification~\cite{marculescu2020edge} on new data thereby making decisions.



However, the traditional approach of deploying centralized ML training on IoT data faces significant obstacles. The training server often grapples with resource limitations, in terms of computational power and storage capacity~\cite{10.1145/3498361.3538917}. Furthermore, the process of transferring vast quantities of data from the edge devices to the central server can introduce significant latency~\cite{pmlr-v151-nguyen22b}. Thus, there is an interest in ``edge'' learning, where the ML models are trained directly on the devices, thus aslo enhancing data privacy.~\cite{marculescu2020edge}. 


Federated learning (FL), a decentralized machine learning approach, has emerged as a compelling method to perform ``edge'' learning. According to McMahan et al.~\cite{mcmahan2017communication}, FL is a decentralized learning paradigm, enabling devices to collaboratively learn a shared model while keeping all the training data on the original device~\cite{mcmahan2017communication}. By shifting the computation closer to the data sources, FL not only reduces latency and bandwidth usage but also enhances data privacy and security.

\begin{figure}[b!]
\centering
\includegraphics[width=0.49\textwidth]{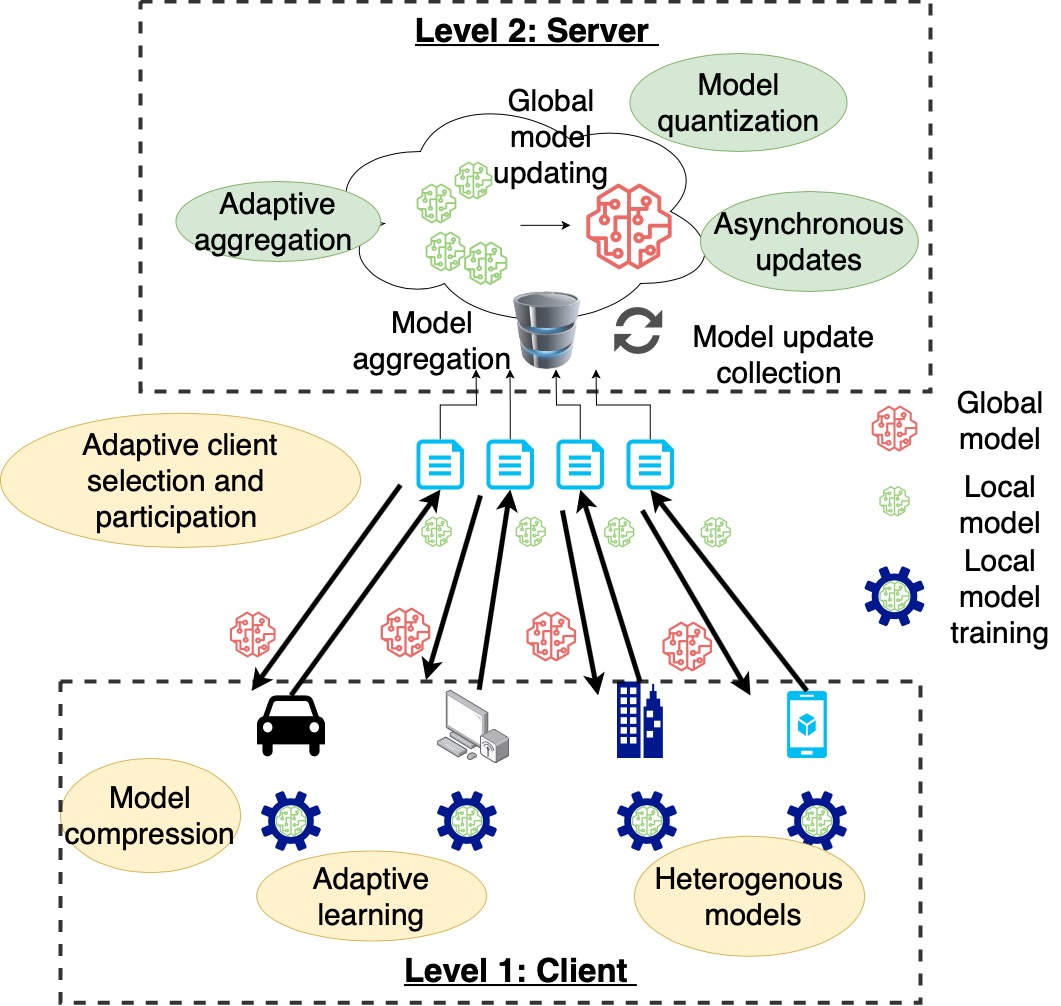}
\caption[List of figure caption goes here]{Overview of the Federated Learning process and the types of optimizations possible on IoT clients (Level 1) and the FL server (Level 2). Yellow bubbles are IoT client optimizations. Green bubbles are FL server optimizations.}
\label{fig:over}
\end{figure}

Most existing FL techniques are designed with the assumption of relatively powerful devices and stable network connections, which is often not the case in IoT environments. Energy efficiency is paramount in this context, as IoT devices often operate on limited power supplies, and excessive energy usage for FL can significantly shorten device lifetimes.
Some challenges of IoT-centric FL systems include the managing the heterogeneous and resource-constrained nature of IoT devices~\cite{marculescu2020edge}, unreliable or unstable networks~\cite{mcmahan2017communication}, scaling to hundreds or thousands of IoT devices~\cite{mcmahan2017communication}, potential delays in model convergence and learning efficiency, and efficient model training and updating in decentralized environments~\cite{10.1145/3437984.3458839}. 
These challenges have opened up a new field of research that focuses on balancing model accuracy and energy efficiency to improve the deployability of FL on low-power IoT devices.


This paper provides a comprehensive review of federated learning in the context of IoT, specifically focusing on resource-constrained devices. We survey existing research and state-of-the-art techniques, addressing the challenges, potential solutions, and future directions in the field. The aim of this survey is to provide a comprehensive review and critical analysis of existing FL techniques within the IoT framework, explicitly considering the unique challenges posed by resource limitations. Recognizing the complex interplay between different components of an FL system, we propose a novel categorization that investigates FL techniques from two perspectives: the client level and the server level. This approach allows us to delve deeper into each component's constraints and requirements, enabling a better understanding of the intricacies involved in deploying FL in IoT devices. 
Fig.~\ref{fig:over} provide an overview of the findings in our survey.

In the following sections, we first highlight prominent IoT client optimizations, and then the FL server optimizations. In the final section, we present our research recommendations and conclude this paper. 
\section{Level 1: Internet-of-Things Client }
\label{sec:compress}

In FL systems, clients locally train model updates on their own data and send these updates to a central server where these updates are aggregated to create a global model reflecting knowledge from all clients~\cite{mcmahan2017communication}. This cycle of local training and global aggregation continues until the model reaches convergence~\cite{mcmahan2017communication}. This decentralized approach, which keeps raw data on client devices rather than a central server, enhancing privacy, is particularly appealing for IoT devices and encourages wider adoption of FL~\cite{marculescu2020edge}.

Training on clients, however, introduces complexity into the system due to device and behavioral heterogeneity, such as differences in computational capabilities, network speeds, and availability for training. Furthermore, clients can be exposed to differing data distributions, leading to a phenomenon known as Non-IID data~\cite{zhao2018federated}, which can pose challenges to the learning process. Therefore, effective client management and optimization are key to maximizing the benefits of federated learning systems. In the next subsections, we provide an overview of some solutions to the aforementioned challenges. We divide the solutions into four broad categories: (1)~Dynamic Client Participation and Selection, (2)~Adaptive Learning, (3)~Model Compression, and (4)~Heterogeneous Models.





\subsection{Dynamic Client Participation and Selection}


Most widely used FL paradigms makes an assumption that all the clients in the FL system are resource sufficient. However, in IoT networks, clients differ significantly~\cite{marculescu2020edge} and can lead to resource wastage, where clients perform training work that does not contribute to enhancing the model, whether due to updates that are ultimately discarded, or poor data distribution. This resource wastage deters users from participating in FL making the scaling to larger deployments problematic.  

To address these challenges, researchers have developed client selection techniques that use new resource-to-accuracy metrics in heterogeneous FL systems. OORT~\cite{lai2021oort} is one such novel FL scheme which accounts for poorly performing clients by assigning each client a trust score. OORT is particularly useful in dealing with resource-constrained FL-based IoT clients because it evaluates each IoT client's resource availability before assigning a task. The FL server constantly monitors the client's activities and updates their trust scores based on their continued performance. This method allows  OORT to improve the time-to-accuracy training time by 1.2$\times-$14.1$\times$ and final model accuracy by 1.3\%$-$9.8\% when compared with the unoptimized FL training scheme.

Despite its advantages, OORT's main downside is the overhead from managing IoT client's trust scores. Subsequent works, such as  PyramidFL~\cite{10.1145/3495243.3517017} and REFL~\cite{10.1145/3552326.3567485}, improve the speed of FL training by designing novel methods to assign representative trust scores. 
In particular, PyramidFL~\cite{10.1145/3495243.3517017} uses the FL server to first determine the initial trust scores and then allows the IoT clients to continually optimize their own trust scores based on the available resources.
In other veins of work, Bonawitz et al.~\cite{10.1145/3133956.3133982} take into account the time of the day to improve the quality of trust scores with temporal data, FedPARL~\cite{Imteaj2021FedPARLCA} accounts for previous training activities to find trustworthy clients, and ~\cite{anelli2019towards} showcases methods to assign trust scores to mobile IoT clients like autonomous vehicles and robotics. FedMCCS~\cite{9212434} assigns priorities to the IoT clients by solving a bi-level optimization problem that considers the availability of resources, communications overhead, and distribution of training data. Another work~\cite{9509751} proposes a dynamic algorithm called ELASTIC based on the trade-off between maximizing the selection and minimizing the energy consumption of the participating clients. ~\cite{9212434,9187798} takes in account communication/network costs and resource capabilities of clients for selection. 



Despite progress in Client Selection optimization, existing literature misses solutions for dynamic, on-the-fly IoT client selection, as most techniques target static environments without client turnover. Potential research areas include enabling clients to share trust scores among each other for optimized participation, integrating selection and participation scoring into overall training, and exploring incentive models in FL for resource-constrained IoT settings.

\subsection{Adaptive Learning}



Adaptive learning involves tailoring the learning process to the specific characteristics, capabilities, or constraints of each client device by adjusting the training parameters. This group of techniques allows for flexible client participation, efficient resource allocation, variable frequency of model updates, and personalized learning approaches for the client IoT devices. 

A common method to perform Adaptive Learning modifies the IoT client's loss function to scale gradient updates based on client characteristics, accommodating heterogeneous client updates. But, deployability challenges arise due to the difficulty of choosing an appropriate loss function modification.


 In order to make better use of system resources and improve performance in a more straightforward manner, FedAwo and FedAwo*~\cite{yu2022federated} use server resources to solve the problems of statistical and system heterogeneity without increasing the load of clients. The authors propose an algorithm for automatic weight optimization (FedAwo), where the server calculates the optimal weight for the local model through the machine-learning algorithm. FedAwo* reduces the training cost by dynamically adjusting the training epoch times of local model training. 
 Another approach to prioritize clients is to modify the number of epochs run on each client. A dynamic epoch parameter in the model training is proposed in BePOCH~\cite{9685095}. Here, an algorithm is used to identify the best number of epochs per training round in an FL model such that the reduces resource consumption and training time. A combination of different Adaptive Learning schemes is yet to be explored, and will be important future work. 
 
 Personalized Federated Learning (PFL) is a group of techniques that create dedicated, resource-efficient local models for individual IoT clients. Techniques such as pFedGate~\cite{chen2023efficient} use a trainable gating layer to speed up training for certain data distributions, and FedSpa~\cite{huang2022achieving} employs personalized sparse masks for custom local models. Despite its success, PFL has drawbacks such as limited adaptability to diverse model architectures and high deployment costs due to substantial server computing needs.

\subsection{Model Compression}

Since most IoT devices have limited memory, model compression is a commonly used technique for reducing memory requirements during the local training phase in FL systems. Compression techniques reduce the number of learnable parameters in the model by (a) factorizing the weight matrices or (b) removing redundant parameters.

\begin{figure}[b!]
\vspace{-0.2in}
\centering
\subfigure[]{\includegraphics[width = 0.49\textwidth]{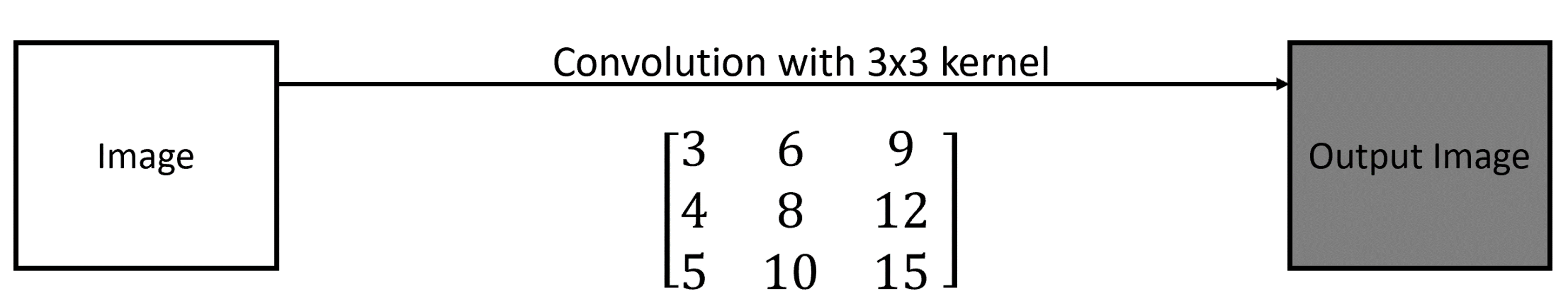}
        }
\subfigure[]{\includegraphics[width = 0.49\textwidth]{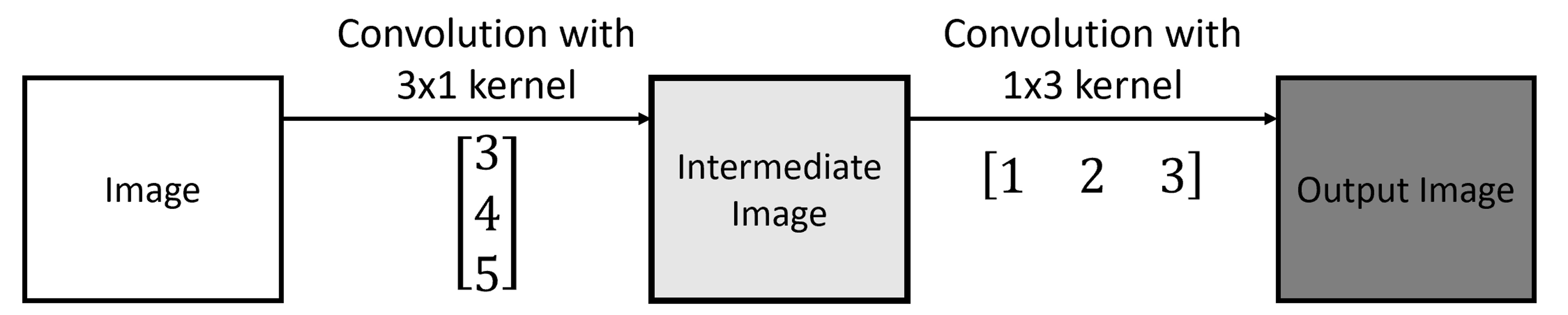}}

\caption{Model compression reduces the \#parameters.}

\label{fig:compression}
\end{figure}

Fig.~\ref{fig:compression} shows an example of weight matrix factorization. 
In this example, a $3 \times 3$ convolution (Fig.~\ref{fig:compression}(a))  kernel is broken up into $3 \times 1$ and $1\times 3$ kernels (Fig.~\ref{fig:compression}(b)). By doing so, the number of parameters decreases from 9 to 6 (33\%).

However, when scaling to larger models, finding such factorizations may not be easy or even possible~\cite{Imteaj2021FedPARLCA, lebedev_speeding-up_2015}. To overcome this problem, researchers~\cite{alvarez_compression-aware_2017, xiong_trp_2020} perform the matrix factorization in a pre-training step. Although matrix factorization reduces the memory requirements, finding optimal matrix factorization may be a time-consuming process. 
Future research should aim to create a theoretical framework for selecting decomposition strategies based on model hyper-parameters. This is crucial for scaling FL as each IoT client might need a distinct decomposition approach.

Other model compression techniques reduce the model memory requirements by identifying and removing redundant parameters~\cite{han_deep_2015}. Research has shown that not all model parameters are needed for the model to converge~\cite{han_deep_2015}. Model pruning techniques have been developed to identify and remove such redundant model parameters~\cite{Imteaj2021FedPARLCA}. These techniques quantify the importance of the parameters by measuring the DNN accuracy losses when the weights are removed. If the accuracy losses are significant, the weight is considered to be important. Similar to factorization, the pruning process is done in a pre-training phase. Pruning techniques are still not used widely because they require special hardware that can perform sparse computation. More research is needed to deploy pruning techniques on general-purpose hardware used in IoT settings.



\vspace{-0.05in}
\subsection{Heterogeneous Models} 


One promising way to enhance FL performance on IoT involves using heterogeneous on-device models. HeteroFL~\cite{diao2021heterofl} was proposed to adaptively assign a subset of global model parameters to an on-device model, assuming the local IoT client model could be subset of a larger server model. However, it's found that not all model architectures can be easily subdivided.

Rather than exchanging gradients or parameters, FedH2L~\cite{li2021fedh2l} shares predictions on a pre-distributed seed set and performs decentralized optimization. Participants focus on finding the best non-conflicting gradient for simultaneously fitting local data and incorporating feedback from peers.


Some recent research has enabled devices to design their IoT client models independently based on federated knowledge distillation techniques~\cite{li2021fedh2l,lin2021ensemble,li2019fedmd}.
In federated knowledge distillation, only the logit information of the FL server is shared with the IoT client models. This enables the IoT clients to combine their local logits with the FL server's logits to update their weights. 
FedZKT~\cite{zhang2022fedzkt} proposes a zero-shot federated distillation approach, contrasting previous research as it requires no on-device data. It allows devices to create models from heterogeneous local resources for knowledge transfer across these models without needing private data.

Transfer Learning methods enhance FL deployability on IoT systems because they help train efficient yet accurate models. For example, Group Knowledge Transfer~\cite{he2020group}, uses an alternating minimization algorithm to train small CNNs for IoT clients, reducing computational requirements and communication while maintaining accuracy.

Another work called hierarchical split federated learning framework~\cite{9964646} efficiently trains FL models through a hierarchical organization of the IoT. This solution reduces the burden on the individual IoT devices by reducing the communication with the distant FL server.

These techniques are successful in scaling FL to large heterogeneous IoT networks but mostly require significant knowledge about every IoT client's hardware resources and/or their private data. Relaxing these constraints through future work would make these solutions more practical.

~\todo{please include below on one lien here} 
\vspace{-0.15in}
\section{Level 2: Federated Learning Server}
\label{sec:compress}

Although the actual learning takes place on individual IoT clients, the server plays a pivotal role to coordinates the learning process, aggregates the model updates, distribute the updated model to different clients, handles device heterogeneity and security
Therefore, the system's overall performance hugely depends on how the server utilizes its resources.


In FL for IoT systems, the model updates are frequently transmitted from multiple devices to the central server. If updates aren't promptly processed and merged, it creates a bottleneck, slowing the learning process. Hence, it's vital to optimize the server's time for aggregating and transmitting updates to IoT devices. Such optimizations involves using parallel computing, smart task scheduling, and enhanced network infrastructure for faster data transmission. Additionally, the server must securely store and manage data from numerous devices without exhausting memory.

In the following subsections, we describe some popular solutions to the aforementioned challenges and highlight some open research areas. For clarity, we divide the solutions into three main categories: (1)~Asynchronous Updates, (2)~Aggregation Algorithms, and (3)~Model Quantization.



\subsection{Asynchronous Updates}

Typically, federated learning systems aggregate updates only after all devices have sent their data. This has been shown to slow down model convergence and increase resource demand~\cite{pmlr-v151-nguyen22b}. It can result in increased communication rounds and idle time for both the client and server.
Sometimes, numerous client updates can overload the server, leading to delays. Many Federated Learning (FL) techniques address this by using asynchronous update mechanisms to optimize server resources and cater to device heterogeneity. Essentially, clients send updates when ready, and servers aggregate them immediately, eliminating wait times for slower devices. The concept of Asynchronous stochastic gradient descent(ASGD)~\cite{dean2012large} is the initial framework that was lead the way for early AsyncFL works~\cite{xie2020asynchronous,chai2021fedat,lian2018asynchronous,9378161}. These techniques decouple model training on the IoT client side from global model by straggler-aware aggregation at the FL server making model converge faster. However, these techniques suffer from a common problem, i.e., data races ~\cite{9853278} on the global model where clients try to update the global model concurrently. 
For example, if multiple clients simultaneously send local models to the server for aggregation, it may result in merging conflicts, leading to inconsistent or incorrect global model updates. Data races on the FL server can lower device utilization and slow training.

To combat the data race problem, some works like FedCrowd~\cite{9853278}, focus on the concurrency of the FL server by using a shadow model at the server along with multiple threads.At a time, one thread aggregates client updates to the shadow model, another copies these updates to the global model, and a final thread dispatches the global model update to clients.
This approach is well suited to tackle the data race problem, however, it has been shown that it cannot be scaled well for production FL systems. This is because the shadow models significantly increase the memory requirements at the FL server. Another approach of using a buffer is discussed in FedBuff~\cite{pmlr-v151-nguyen22b}. Here, the buffer at the FL server is used to store the client updates and the FL server uses this buffer to aggregate and update the global model. FedBuff reports to be 2.5x more efficient than the basic FedAsync. This work extends to PAPAYA~\cite{huba2022papaya} which is a production-ready secure aggregation asynchronous protocol.

Another important consideration for Asynchronous Updates is the variance in the processing speeds among the IoT clients. Faster devices participate in global updates many more times than slow devices, and some slow devices cannot join in the global aggregation even once due to staleness control.
The stale model problem can come up in the aforementioned approaches, where the newly arrived update is calculated based on a set of stale weights that are older than the current global model, thus hurting the convergence of the model. AsyncFedED~\cite{wang2022asyncfeded} and TimelyFL~\cite{zhang2023timelyfl} present asynchronous federated learning frameworks with an adaptive weight aggregation algorithm to solve the staleness issue with heterogeneous clients. 
However, this problem is far from solved, and more research is required to ensure the resources in large heterogeneous IoT networks are being maximally utilized. Furthermore, new training algorithms that focus on optimizing time-to-accuracy metrics are required.
~\todo{Abhinav: please add future work direction and conclude this section}

\subsection{Aggregation Algorithms }

In FL, clients send their updates which can be in the form of gradients, parameter differences, or even entire model weights, and the server uses an aggregation method to combine these. Usually, the server uses averaging~\cite{mcmahan2017communication}, where the server calculates the average of the updates received from the device. The choice of the aggregation algorithms usually does not change the computational complexity of the FL system, but can significantly impact the number of iterations required to reach convergence.~\cite{Jiang_2023}. Reducing the number of training iterations can save energy on the server as well as the participating IoT clients. These techniques also dramatically reduce communication overheads.


Concept of Hierarchical Aggregation~\cite{liu2019clientedgecloud,9054634,Wu_2020} has been introduced where aggregation model updates from clients are aggregated at multiple levels of the hierarchy. It helps to relieve communication overhead and reduce bandwidth requirements while maintaining privacy and improving the scalability of the system. Similarly motivated research~\cite{Wu_2020,liu2019clientedgecloud,9054634} describes methods that use a three-layer hierarchy of Mobile Edge Computing. 
This design approach is suitable in FL systems where plenty of resources are available in the IoT clients (because of large numbers) thus allowing the offloading of work from the server. 
~\todo{Abhinav: please correct the following sentence english}
Aggregation algorithms that dynamically adapt the frequency of global aggregation under a fixed resource budget have also been developed~\cite{8664630}. In such techniques, a controller learns the data distribution, system dynamics, and model characteristics to ensure that the most important model updates are processed first. Reducing the number of model updates also decreases the memory and computation requirements~\cite{Ghosh2020AnEF}.



Researchers have found that the standard aggregation algorithms such as FedAvg~\cite{mcmahan2017communication} take several times longer to converge on non-IID data. There has been a recent focus on adaptive aggregation methods based on data heterogeneity across IoT clients. Some works consider adding momentum-based optimizers such as SLOWMO~\cite{wang2020slowmo} and FedAdam~\cite{reddi2020adaptive} to improve the convergence time, and consequently the computational burden on the FL server. ~\cite{huang2021behavior} propose an attention-based aggregation technique that combines each IoT client's data distribution with the data distribution of the entire group. Although these techniques show promising results, they are not able to completely avoid the deterioration of model convergence due to the client drift caused by non-IID local updates. This is an important area of future research.

\begin{figure}[b!]
\vspace{-0.2in}
\centering
\includegraphics[width=0.4\textwidth]{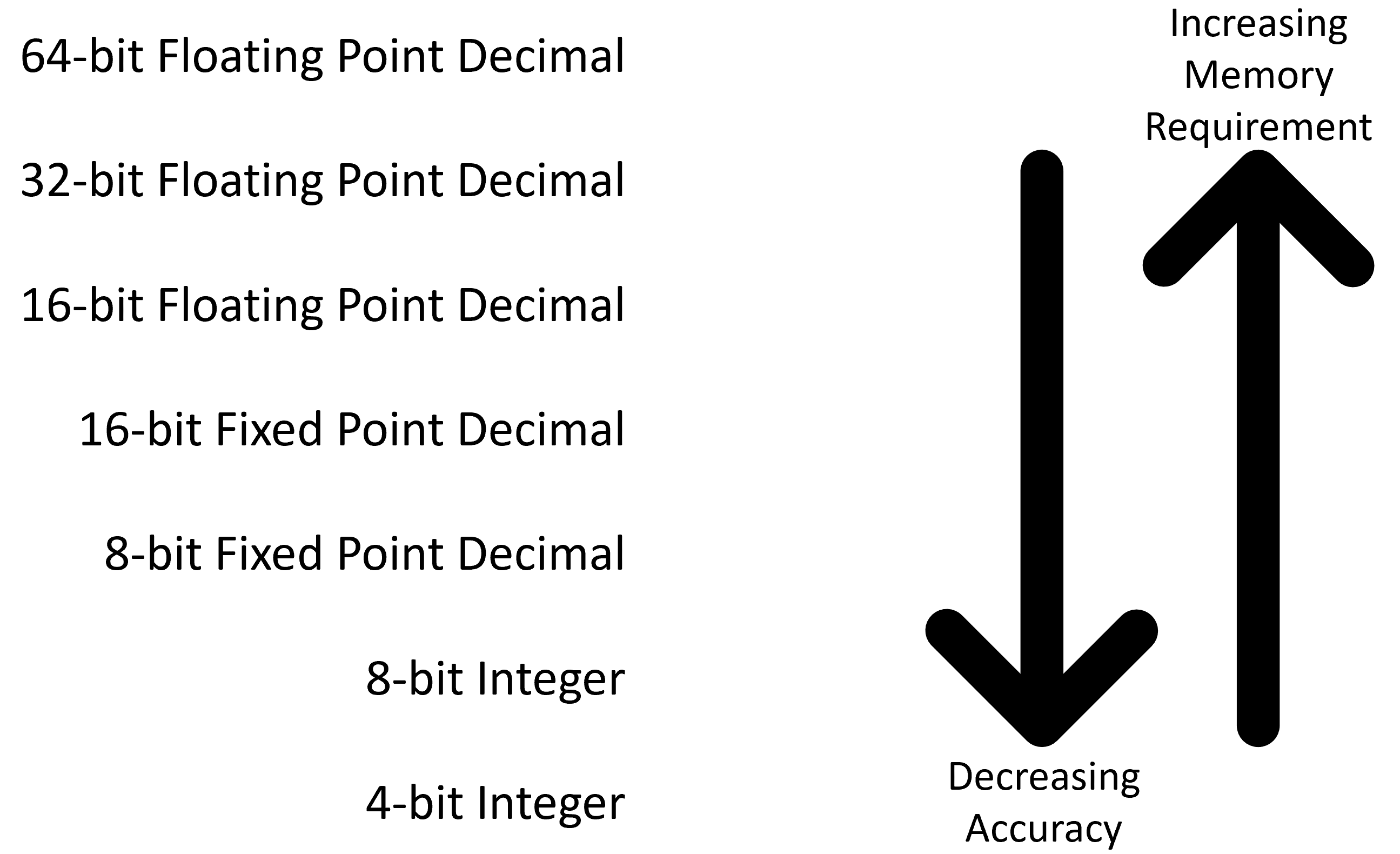}
\caption[List of figure caption goes here]{Decreasing the precision of the model reduces the memory requirement, but also impacts the accuracy.}
\label{fig:quant}
\end{figure}

\subsection{Model Quantization}

By default, most numbers in machine learning libraries are represent floating-point values using 32 bits~\cite{krishnamoorthi_quantizing_2018}. These 32-bit values are not always needed during the training phase~\cite{han_deep_2015}. There are numerous benefits to using formats with lower precision than 32-bit floating point: (1)~requires less memory, enables the FL server to hold larger models and coordinate more client updates. (2)~requires less memory bandwidth which speeds up data transfer operations~\cite{9305988}. (3)~math operations run faster in reduced precision~\cite{9054168}.

However, there exists a tradeoff when using extremely low-precision representations. Fig.~\ref{fig:quant} highlights this tradeoff. When using 4-bit or 8-bit formats, it is important to consider training convergence. The model may not converge if all the operations are performed in the 4-bit format~\cite{10.1145/3437984.3458839}. Thus, it is an open research problem to automatically identify which precision is suitable for the different model parameters. Solving this problem would enable more effective scaling of FL systems. Enhancing software support to support non-standard quantization formats is another important area of open research.




\section{Discussion and Conclusion}
\label{sec:compress}

In this section, we propose metrics that will help researchers evaluate their solutions' deployability on IoT FL systems. We summarize our findings (in TABLE~\ref{tab:comp}) and conclude this paper.

\begin{table}[b!]
\vspace{-0.1in}
\centering
\begin{tabular}{|C{2cm}|L{5.8cm}|}
\hline
Application Area & \multicolumn{1}{c|}{Open Research Areas} \\ \hline
IoT Client & High accuracy with small models; consistent model convergence with non-IID data.   \\ \hline
FL Server & Reduce accuracy losses with int4 weights; improve convergence rates asynchronous updates. \\ \hline
\end{tabular}
\caption{Open research areas to improve the deployability of FL on IoT devices. }
\label{tab:comp}
\end{table}


\subsection{Proposed Evaluation Metrics}
\begin{enumerate}
    \item Non-IID Datasets: The FL papers mostly use synthetic Non-IID~\cite{zhao2018federated,wang2022generative} split techniques on datasets like CIFAR-10 and ImageNet that are not representative of real-world IoT data~\cite{10.1145/3498361.3538917, pmlr-v151-nguyen22b}. The data from different devices may have different temporal and data distributions, lighting conditions, and angles. Testing on datasets such as Market-1501 or having a benchmark technique of generating synthetic non-IID split may be more valuable. 

    \item Baselines for time-to-accuracy: FL techniques are often measured on the amount of time taken to converge for certain accuracy on a dataset. However, most techniques are measured on different baselines, e.g., FedBuff~\cite{pmlr-v151-nguyen22b} presents results for the time taken to reach 60\% accuracy on CIFAR-10, while FedBalancer~\cite{10.1145/3498361.3538917} measures the time taken to reach 80\% on FEMNIST. Standardizing the baselines is important for future research.


    \item Energy Delay Product (EDP): EDP combines energy consumption and latency by multiplying the energy consumed by the FL model with the corresponding latency. This is important for both the IoT clients and the FL servers to ensure that neither is becoming a bottleneck.

    \item Round efficiency: The number of communication rounds taken by the system to converge helps measure the performance of the aggregation algorithms.
    
\end{enumerate}

\subsection{Conclusion}

FL systems have become more ubiquitous in the last few years because of the widespread deployment of IoT devices. FL is well suited to train large-scale machine-learning models because of its ability to guarantee convergence, preserve privacy, and save network bandwidth. However, deploying FL in IoT environments has many challenges: (1)~large numbers of FL clients, (2)~unreliable or unstable networks, (3)~heterogeneous clients, and (4)~limited computing resources and memory. In this survey, we highlight and summarize the key research topics that help improve the deployability of FL in IoT environments. We find that the existing research can be divided into optimizations for the FL server and the IoT clients. Within each type of optimization, we find that similar solutions can be categorized together. We use this categorization to highlight the advantages, disadvantages, and potential future work. We also provide insights about evaluation metrics and how they can be improved.


\renewcommand*{\bibfont}{\footnotesize}
\printbibliography

\end{document}